\documentclass[11pt]{article}

\usepackage[final]{acl}

\usepackage{times}
\usepackage{latexsym}
\usepackage{booktabs}
\usepackage{tabularx}
\usepackage{multirow}
\usepackage{booktabs}
\usepackage{pifont}
\usepackage{enumitem}
\usepackage[T1]{fontenc}

\usepackage[utf8]{inputenc}

\usepackage{microtype}

\usepackage{inconsolata}

\usepackage{listings}
\usepackage{xcolor}

\usepackage{graphicx}
\usepackage{xspace}
\usepackage{xcolor}
\usepackage{enumitem}
\usepackage{colortbl} 
\usepackage[most]{tcolorbox}
\tcbuselibrary{skins,breakable}
\usepackage{arabtex}
\usepackage{tabularx}
\usepackage{multicol}
\usepackage{fancyvrb} 
\usepackage{subcaption}
\usepackage{utf8}
\setcode{utf8} 
\newcommand{\dataset}{\textsc{IslamicFaithQA}\xspace}

\newtcolorbox{QABlock}[1]{%
  enhanced,
  width=\columnwidth,
  colback=white,
  colframe=black!60,
  boxrule=0.4pt,
  arc=1.2mm,
  left=1.2mm,right=1.2mm,top=1.0mm,bottom=1.0mm,
  boxsep=0.8mm,
  fontupper=\footnotesize,
  before skip=0.6\baselineskip,
  after skip=0.6\baselineskip,
  title=\textbf{#1},
  fonttitle=\footnotesize\bfseries,
  coltitle=black,
  attach boxed title to top left={yshift=-1.2mm, xshift=1.2mm},
  boxed title style={
    colback=white,
    colframe=white,   %
    boxrule=0pt,
    sharp corners,
    left=0pt,right=0pt,top=0pt,bottom=0pt
  },
  segmentation style={draw=none}
}
\newcommand{\QuranCite}[1]{%
  \begingroup
  \setlength{\fboxsep}{1.2pt}%
  \colorbox{blue!8}{\textbf{Reference - Qur'an: #1}}%
  \endgroup
}
\newcommand{\EvalTag}[1]{%
  \begingroup
  \setlength{\fboxsep}{1.2pt}%
  \ifnum\pdfstrcmp{#1}{Correct}=0
    \colorbox{green!15}{\textbf{Judge: #1}}%
  \else
    \colorbox{red!12}{\textbf{Judge: #1}}%
  \fi
  \endgroup
}
\usepackage{multicol}
\usepackage{listings}

\usepackage{graphicx}
\usepackage{tabularx}
\usepackage{xcolor} %
\definecolor{lightblue}{rgb}{.50,.90,0.51}
\definecolor{tri}{rgb}{.25,.88,.82}
\definecolor{lilac}{rgb}{0.85,0.64,0.85}
\definecolor{atomictangerine}{rgb}{1.0, 0.6, 0.4}

\title{From RAG to Agentic RAG for Faithful Islamic Question Answering}

\author{
Gagan Bhatia$^{1}$,
Hamdy Mubarak$^{1}$,
Mustafa Jarrar$^{2}$,
George Mikros$^{2}$,\\
{\bf
Fadi Zaraket$^{3}$,
Mahmoud Alhirthani$^{2}$,
Mutaz Al-Khatib$^{4}$,
}\\
{\bf
Logan Cochrane$^{5}$,
Kareem Darwish$^{1}$,
Rashid Yahiaoui$^{2}$,
Firoj Alam$^{1}$
}\\
$^{1}$ Qatar Computing Research Institute, HBKU, Qatar, \\
$^{2}$ College of Humanities and Social Sciences, HBKU, Qatar, 
$^{3}$ Arab Center for Research \\ and Policy Studies, Qatar,
$^{4}$ College of Islamic Studies, HBKU, Qatar \\
$^{5}$ College of Public Policy, HBKU, Qatar \\
\texttt{fialam@hbku.edu.qa}
\\ \small\href{https://huggingface.co/collections/QCRI/islamic-knowledge-in-llms}{https://huggingface.co/collections/QCRI/islamic-knowledge-in-llms}
}

\begin{document}
\maketitle
\begin{abstract}
Large Language Models (LLMs) are increasingly used for Islamic question answering, where ungrounded responses may carry serious religious consequences. Yet standard MCQ/MRC-style evaluations\footnote{MCQ: Multiple choice questions, MRC: Machine Reading Comprehension} do not capture key real-world failure modes, notably free-form hallucinations and the ability to abstain when evidence is insufficient. To address this gap, we introduce \dataset, a 3{,}810-item bilingual (Arabic/English) \emph{generative} benchmark with atomic single-gold answers, which enables direct measurement of hallucination and abstention. We additionally developed an end-to-end grounded Islamic modeling suite consisting of \textit{(i)} 25K Arabic text-grounded SFT reasoning pairs, \textit{(ii)} 5K bilingual preference samples for reward-guided alignment, and \textit{(iii)} a verse-level Qur'an retrieval corpus of $\sim$6k atomic \textit{verses} (ayat). Building on these resources, we develop an agentic Quran-grounding framework (agentic RAG) that uses structured tool calls for iterative evidence seeking and answer revision. Experiments across Arabic-centric and multilingual LLMs show that retrieval improves correctness and that agentic RAG yields the largest gains beyond standard RAG, achieving state-of-the-art performance and stronger Arabic--English robustness even with a small model (i.e., Qwen3 4B). We made the datasets are publicly available.\footnote{\href{https://huggingface.co/datasets/QCRI/IslamicFaithQA}{https://huggingface.co/datasets/QCRI/IslamicFaithQA}}
\end{list}
\end{abstract}

\section{Introduction}
Large language models (LLMs) are increasingly positioned as general-purpose assistants for decision support, education, and guidance in value-laden domains. However, a persistent challenge is that fluent generations can obscure \emph{normative} and \emph{factual} unreliability: models remain sensitive to framing, role instructions, %
and they may produce confident but unsupported responses \citep{jiao2025llm}. 

\begin{figure}[t]
    \centering
    \includegraphics[width=\linewidth]{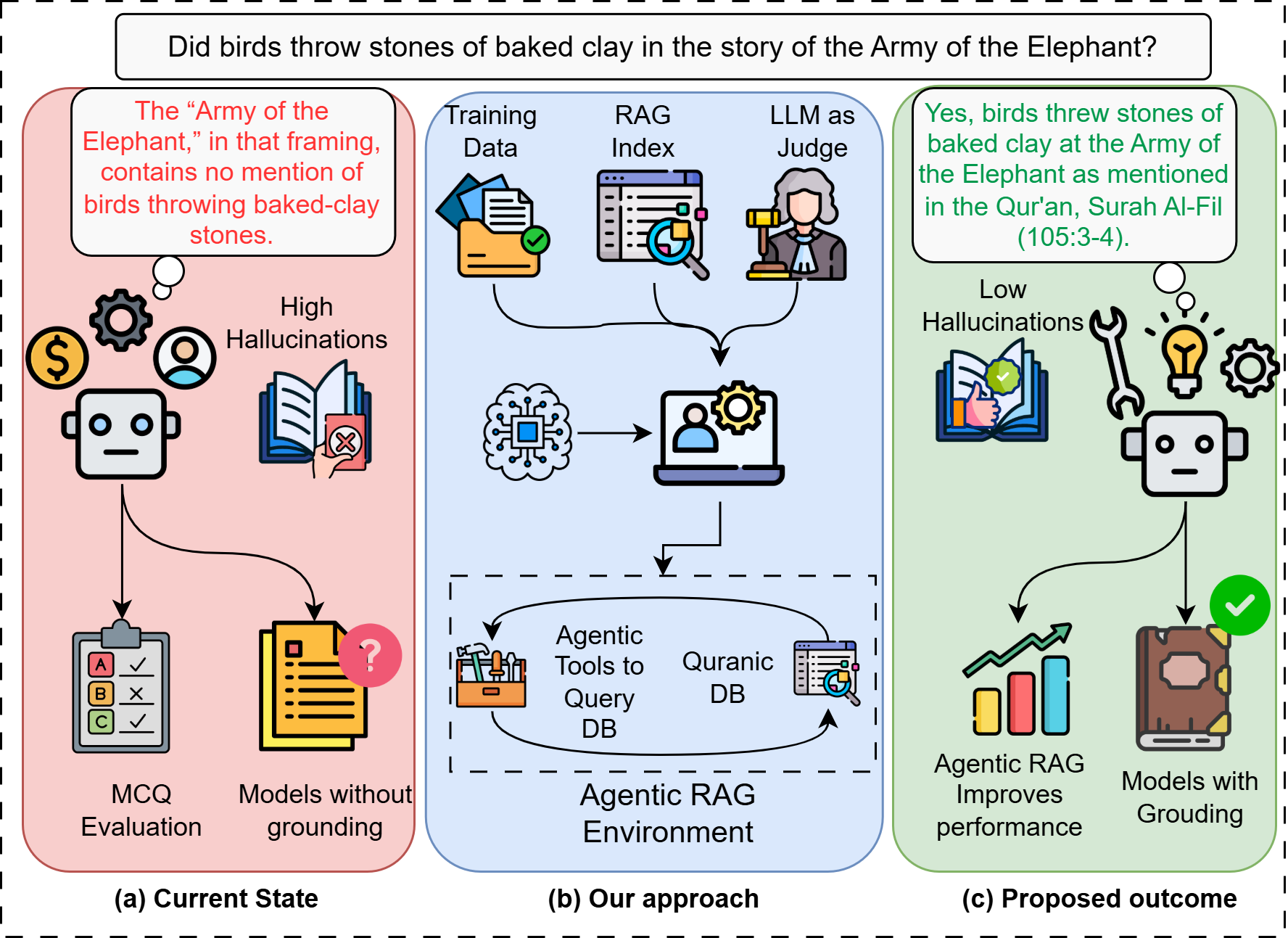}
    \vspace{-0.2cm}
    \caption{\textbf{Current--Proposed--Outcome.}
    \textit{(a)} Current Islamic QA.
    \textit{(b)} We combine \dataset{}, LLM judging, Quran retrieval, and agentic evidence seeking.
    \textit{(c)} This yields more faithful, citation-backed 
    responses.}    

    \label{fig:islamicsimpleqa-framework}
    \vspace{-0.3cm}
\end{figure}

Islamic question answering is a particularly challenging testbed for \textit{reliability} problem. Deployed Islamic QA systems\footnote{e.g., \url{https://ansari.chat/}, \url{https://usul.ai}, \url{https://wisqu.ai}} indicate strong demand, yet their proprietary evaluations highlight the need for shared benchmarks that emphasize grounding, citation fidelity, and abstention. Unlike general information-seeking queries, Islamic QA is embedded in jurisprudential reasoning (\emph{fiqh}), school-of-thought conventions, and culturally situated norms that demand faithful grounding in canonical sources and careful handling of uncertainty. 
Recent multilingual and culture-aware evaluations show that moral judgments and alignment behaviour vary meaningfully with language and data provenance, with persistent representational bias and Western-dominance effects that are especially salient for non-Western normative systems \citep{naous2025on,guo2025care}. 
Within the Islamic domain, emerging resources (e.g., inheritance-law reasoning and abstention-aware fiqh evaluations) indicate both progress and substantial performance gaps, particularly for Arabic and for school-aware nuance, reinforcing the need for fine-grained reliability checks tailored to Islamic jurisprudence \cite{bouchekif2025assessing,elsafoury2025out,asseri2025prompt}. Parallel work on Quranic retrieval-augmented generation (RAG) further suggests that grounding can improve faithfulness, but that outcomes are mixed and depend on model capacity and retrieval quality \citep{khalila2025investigating}.

A central obstacle in knowledge-intensive Islamic QA is \textit{hallucination}. Multilingual studies suggest that Arabic settings can amplify factuality and faithfulness errors, and that coarse answer-level metrics often miss subtle inconsistencies important for normative argumentation \citep{islam2025how,alansari2025arahallueval,hosseini2025perhallueval,elchafei2025span-level,wang2025joint}. Moreover, test-time scaling results show that longer reasoning traces do not reliably improve grounding and may even increase overconfident errors \citep{gema2025inverse,zhao2025test-time}. This motivates retrieval-based grounding, especially agentic setups that interleave search, tool use, and verification, but practical reliability depends on robust tool orchestration and domain ontologies \citep{liang2025reasoning}. Accordingly, we target three under-specified and under-measured needs in Islamic QA: \textit{(i)} Arabic--English robustness, \textit{(ii)} calibrated abstention under insufficient evidence, and \textit{(iii)} evidence-grounded generation aligned with canonical sources \citep{bhatia-etal-2025-swan}.
Figure~\ref{fig:islamicsimpleqa-framework} summarizes our motivation and method in a current--proposed--outcome view, contrasting today’s Islamic QA pipeline with our Islamic grounding-based approach and its resulting citation-backed bilingual answers. 

Our contributions are as follows:
\begin{itemize}[noitemsep,topsep=0pt,leftmargin=*,labelsep=.5em]
    \item \textbf{Bilingual Islamic QA benchmark:} \dataset\ comprises 3{,}810 Arabic--English questions with atomic, single-gold answers\footnote{Many Islamic questions \emph{allow} multiple valid answers across interpretive traditions (madh\=ahib). To enable reliable generative evaluation, we focus on \emph{atomic} items with a single text-grounded answer; handling disputed cases via multi-reference/equivalence-class grading is left to future work.} and a strict \texttt{correct}, \texttt{incorrect} or \texttt{not attempted} labeling scheme, enabling direct measurement of \textit{hallucination} and \textit{abstention}.
    
    \item \textbf{An end-to-end data suite for grounded Islamic modeling:} We release a unified set of resources spanning \textbf{25K} Arabic text-grounded SFT reasoning pairs, \textbf{5K} bilingual preference samples for reward-guided alignment, and a verse-level Quran retrieval corpus of \textbf{6,236} atomic \textit{ayat}.
    \item \textbf{Evidence-seeking inference via agentic Quran grounding:} We develop and evaluate an \emph{agentic RAG} setup that turns retrieval into an explicit decision process through structured tool calls (semantic search, verse reading, metadata lookup).
\end{itemize}

Across all backbones, \dataset{} exposes a substantial reliability gap between general instruction-following fluency and text-grounded Islamic correctness. Most off-the-shelf multilingual LLMs remain below $30\%$ accuracy under strict LLM-as-Judge grading (Table~\ref{tab:leaderboard}). 
Retrieval augmentation is the most consistently effective intervention, improving performance across models by anchoring generations to canonical evidence (Table~\ref{tab:ablation}). Most notably, \emph{agentic} RAG yields the largest gains beyond standard RAG, enabling strong bilingual robustness by forcing iterative evidence seeking and verse inspection before answering. For Qwen3-4B-2507, accuracy improves from $21.85$ (base) to $38.85$ (+RAG) and to $48.90$ (+Agentic RAG), while also narrowing the Arabic--English gap (Table~\ref{tab:ablation}). Finally, combining a strong in-domain backbone with agentic grounding achieves the best overall performance, with Fanar-2-27B + Agentic RAG reaching $57.30$ average accuracy (Table~\ref{tab:ablation}).

\section{Related Work}
\subsection{Benchmarking in Islamic Domain}
General-purpose evaluations of moral and trustworthiness show that LLM behavior is highly sensitive to framing and may appear competent while remaining unreliable, motivating the need for domain-grounded assessment beyond generic scenarios \citep{alkhalifa2025allms,abhishek2025beats,hassan-bhatti-etal-2025-cultranai,larabench}. Follow-on work in specialized, high-stakes settings (e.g., legal/medical ethics) emphasizes stricter correctness notions, risk-aware protocols, and evaluation designs that better reflect real deployment constraints \citep{shao2025when,hong2025towards,jin2025medethiceval,hui2025trident}. In culturally situated contexts, multilingual studies further show that moral judgments and alignment behavior vary substantially with language and data provenance, with recurring Western-dominance effects and representational bias \citep{naous2025on,guo2025care,agarwal2024ethical}. Within Islamic QA specifically, recent benchmarks and datasets begin to target fiqh-style reasoning, abstention, and culturally faithful evaluation, however, consistently report gaps in Arabic performance and jurisprudential nuance \citep{atif2025sacred,bouchekif-etal-2025-qias,lahmar2025islamtrust,mubarak-etal-2025-islamiceval,elsafoury2025out,aljaji2025quranBenchmark,palmx2025,tajrin-etal-2025-aya}.
These limitations motivate our focus on \emph{open-ended} Islamic QA with \emph{atomic single-gold} answers and strict LLM-as-a-judge grading to directly measure hallucination and abstention, rather than relying on MCQ/MRC-style evaluations \citep{haas2025simpleqaverifiedreliablefactuality}.

\subsection{Knowledge-Intensive Domains}
Hallucination remains a central failure mode in knowledge-intensive QA. Recent multilingual and Arabic-focused studies report elevated factuality and faithfulness errors, and call for evaluation beyond answer-only metrics, including span-level attribution and joint assessment of reasoning traces and final outputs \citep{islam2025how,alansari2025arahallueval,elchafei2025span-level,wang2025joint}. At the same time, evidence from test-time scaling shows that longer reasoning traces do not reliably improve grounding and can increase overconfident errors, reinforcing that ``thinking more'' is not a substitute for evidence \citep{gema2025inverse,zhao2025test-time}. Retrieval augmentation is therefore a key mechanism for improving groundedness. Prior work on reasoning and agentic RAG highlights that iterative search, tool use, and verification can improve faithfulness when supported by reliable retrieval and orchestration \citep{li2025towards}. In Qur’anic/Islamic settings, empirical studies shows that RAG can improve faithfulness, although performance depends strongly on retrieval quality, model capacity, and domain coverage \citep{khalila2025investigating,salameh2024quranicAudio}. Broader trustworthiness suites emphasize that factuality should be assessed alongside safety and misinformation risk in value-laden deployments \citep{huang2023trustgpt,abhishek2025beats,hui2025trident}, while Arabic-centric resources further highlight how language coverage and representation affect retrieval and downstream reliability \citep{bhatia-etal-2025-swan}. These findings motivate our comparison of standard RAG versus \emph{agentic} RAG under a strict generative, abstention-aware protocol designed to 
reduce hallucinations in Islamic QA \citep{haas2025simpleqaverifiedreliablefactuality,abbas2026fanarsadiq}.

\section{Datasets}
\label{sec:datasets}

To facilitate the development of robust Islamic LLMs and enable precise hallucination evaluation, we construct a comprehensive suite of resources comprising instruction tuning data, preference alignment data, a retrieval corpus, and a novel evaluation benchmark, \dataset. The specific statistics for each set of our data suite are summarized in Table~\ref{tab:data_stats}.

\begin{table}[h]
\centering
\small
\renewcommand{\arraystretch}{1.2}
\resizebox{\columnwidth}{!}{
\begin{tabular}{llrl}
\toprule
\textbf{Dataset} & \textbf{Role} & \textbf{Size} & \textbf{Language} \\
\midrule
SFT Reasoning & Training & 25,000 & Arabic \\
RL Preference & Training & 5,000 & Ar + En \\
Quran RAG & Retrieval & 6,236 & Arabic \\
\textbf{\dataset} & \textbf{Evaluation} & \textbf{3,810} & \textbf{Ar + En} \\
\bottomrule
\end{tabular}}
\vspace{-0.2cm}
\caption{Summary of the constructed data resources. Sizes represent the number of instruction pairs, reward samples, or atomic retrieval units (verses).}
\label{tab:data_stats}
\vspace{-0.4cm}
\end{table}

\subsection{Training and Alignment Resources}

We develop two training datasets and a Quranic RAG Index to enhance model capability in the Islamic domain, specifically targeting theological reasoning and safety alignment. 

\noindent
\paragraph{SFT Reasoning Dataset.}
For Supervised Fine-Tuning (SFT), we curate a dataset of 25,000 bilingual samples (Arabic and English) instruction-response
pairs centered on theological reasoning. Unlike standard QA pairs, this dataset is text-grounded; questions are derived directly from Quranic verses and Hadith, with answers requiring grounded reasoning steps rather than simple extraction. \textcolor{black}{As shown in Figure~\ref{fig:pipeline} we use LLM generated datasets.} This structure facilitates the model's ability to articulate the logical basis behind Islamic rulings. An example of the SFT Reasoning dataset is given in Appendix~\ref{app:sft_eg}.

\noindent
\paragraph{RL Preference Dataset.}
To support preference optimization techniques such as GRPO \cite{shao2024deepseekmathpushinglimitsmathematical}, we construct a Reinforcement Learning (RL) dataset of 5,000 bilingual samples (Arabic and English). Each instance includes a question derived from canonical texts, a gold-standard answer, and specific evaluation parameters designed to train reward models. This dataset is crucial for aligning model outputs with factual correctness and minimizing hallucination in sensitive religious contexts. An example 
is provided in Appendix~\ref{app:rl_eg}.

\noindent
\paragraph{Quran RAG Dataset.}
Additionally, for 
RAG experiments, we process the standard corpus of the Holy Quran into 6,236 retrieval units corresponding to individual \textit{Ayat} (verses), serving as the ground-truth knowledge base for both generation and evaluation tasks. Concretely, we segment the full Qur'an into 6,236 units (one \textit{ayah} per record) and attach standardised metadata required for tool use and evaluation, including \textit{surah} and \textit{ayah} indices, canonical verse identifiers, and normalised Arabic text (to reduce orthographic variance and improve dense retrieval). This structure enables \textit{(i)} consistent verse-level citation in model outputs, \textit{(ii)} deterministic mapping from retrieved evidence to a unique canonical reference, and \textit{(iii)} faithful evaluation of grounding by checking whether predicted claims are supported by retrieved \textit{ayat}.

\subsection{The \dataset Benchmark}

Existing evaluations for Islamic NLP often rely on discriminative formats like MCQ  \cite{palmx2025, bouchekif-etal-2025-qias} or 
MRC \cite{bashir2021arabic}. As detailed in Table~\ref{tab:comparison}, these formats allow models to guess correctly without genuine grounding and fail to measure \textit{abstention} capabilities. To address this, we introduce \dataset, a bilingual generative benchmark with \textbf{3,810} Arabic questions and English questions, designed to measure hallucination rates via an LLM-as-a-Judge protocol.

\begin{table}[h]
\centering
\setlength{\tabcolsep}{3.0pt}
\renewcommand{\arraystretch}{1.0}
\resizebox{\columnwidth}{!}{
\begin{tabular}{llccccc}
\toprule
\textbf{Resource} & \textbf{Type} & \textbf{Size} &
\textbf{EN+AR} &
\textbf{Text-} &
\textbf{Format} &
\textbf{GenQA} \\
& & & & \textbf{grounded} & & \\
\midrule
\rowcolor{gray!10}
\textbf{\dataset (Ours)} & \textbf{Benchmark} & \textbf{3,810} &
\ding{51} & \ding{51} & \textbf{GenQA} &
\ding{51} \\
\midrule
QRCD \cite{bashir2021arabic} & Dataset & 1,337 &
\ding{55} & \ding{51} & MRC &
\ding{55} \\
AyaTEC \cite{malhas2020ayatec} & Dataset & 207 &
\ding{55} & \ding{51} & VerseQA &
\ding{55} \\
Hajj-FQA \cite{aleid2025hajj} & Dataset & 2,826 &
\ding{55} & \ding{55} & FatwaQA &
\ding{55} \\
IslamTrust \cite{lahmar2025islamtrust} & Benchmark & 406 &
\ding{51} & \ding{55} & MCQ &
\ding{55} \\
\midrule
Qur'an QA 2022 \cite{malhas-etal-2022-quran} & Shared task & 1,337 &
\ding{55} & \ding{51} & MRC &
\ding{55} \\
IslamicEval 2025 \cite{mubarak-etal-2025-islamiceval} & Shared task & 1,506 &
\ding{55} & \ding{51} & PR &
\ding{55} \\
QIAS 2025 \cite{bouchekif-etal-2025-qias} & Shared task & 22,000 &
\ding{55} & \ding{51} & MCQ &
\ding{55} \\
PalmX 2025 \cite{palmx2025} & Shared task & 1,900 &
\ding{55} & \ding{55} & MCQ &
\ding{55} \\
\bottomrule
\end{tabular}
}
\caption{
Comparison of \dataset with prominent Islamic NLP resources.
\textbf{Size} reports the primary evaluation unit (e.g., QA pairs / MCQs; for IslamicEval it is annotated answers).
\textbf{Text-grounded} denotes questions grounded in canonical texts.
\textbf{Format:} GenQA = Generative QA; MRC = Machine Reading Comprehension; PR = Passage Retrieval; MCQ = Multiple Choice. 
}
\label{tab:comparison}
\end{table}

\subsubsection{\dataset Curation Pipeline}
\begin{figure}[!t]
    \centering
    \includegraphics[width=\columnwidth]{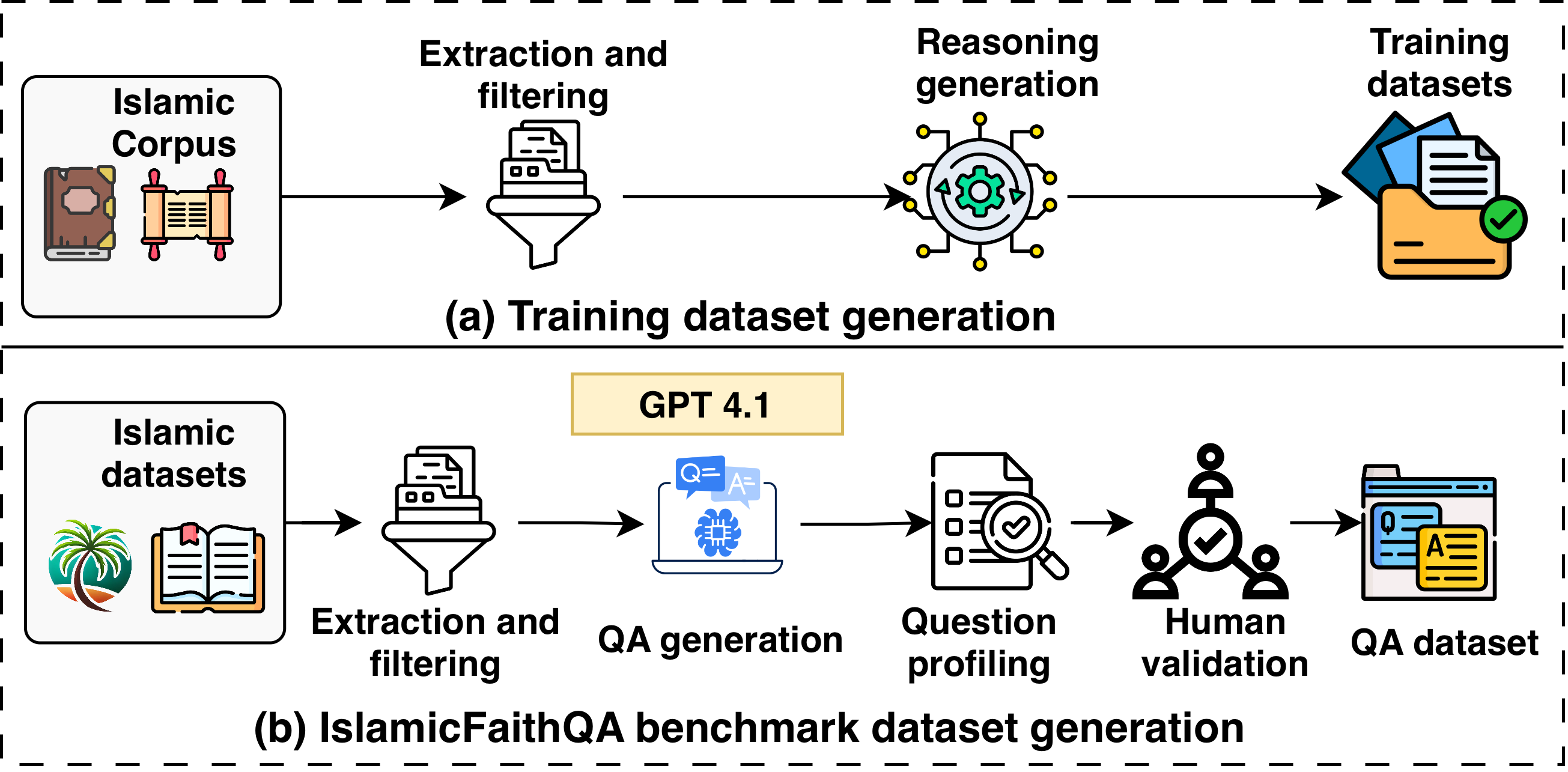}
    \caption{The construction pipeline for \dataset. 
    }
    \label{fig:pipeline}
\end{figure}

As illustrated in Figure~\ref{fig:pipeline}, we employ a rigorous semi-automated pipeline. We aggregate high-quality samples from sources such as Hajj-FQA \cite{aleid2025hajj}, QIAS \cite{bouchekif-etal-2025-qias}, and PalmX \cite{palmx2025}. 

\paragraph{Extraction and Filtering.} In addition to QAs some datasets also include difficulty label as a metadata. In this step, we select QAs from datasets that include difficulty annotations, we retain only the hardest examples. 

\paragraph{QA Generation.} The selected questions are then reformulated by GPT-4.1 into short, fact-based generative questions with atomic gold answers.

\paragraph{Question Profiling.} To enrich the benchmark, we add a layer of question-level metadata through an additional profiling step. We use a separate LLM, GPT-4.1, as an expert annotator to assign \textit{(i)} a difficulty level on a five-point scale ranging from \textit{``very easy'' (score 1)} to \textit{``very hard'' (score 5)}; \textit{(ii)} a binary label indicating whether the question requires reasoning; \textit{(iii)} a binary label indicating whether answering the question requires multi-step reasoning; and \textit{(iv)} a single fine-grained question category from a fixed taxonomy: inheritance law, jurisprudence, prophetic biography, Islamic creed, Qur'anic studies, hadith studies, Islamic finance and economics, Islamic ethics and morality, Islamic history, Islamic family law, contemporary issues, and comparative religion.

The prompts for \textit{QA generation} and \textit{question profiling} are provided in Appendix~\ref{app:data_prompts} and Appendix~\ref{app:data_prompt_diff}, respectively.

\begin{figure}[t]
    \centering
    \begin{subfigure}[b]{0.4\textwidth}
        \includegraphics[width=0.98\textwidth]{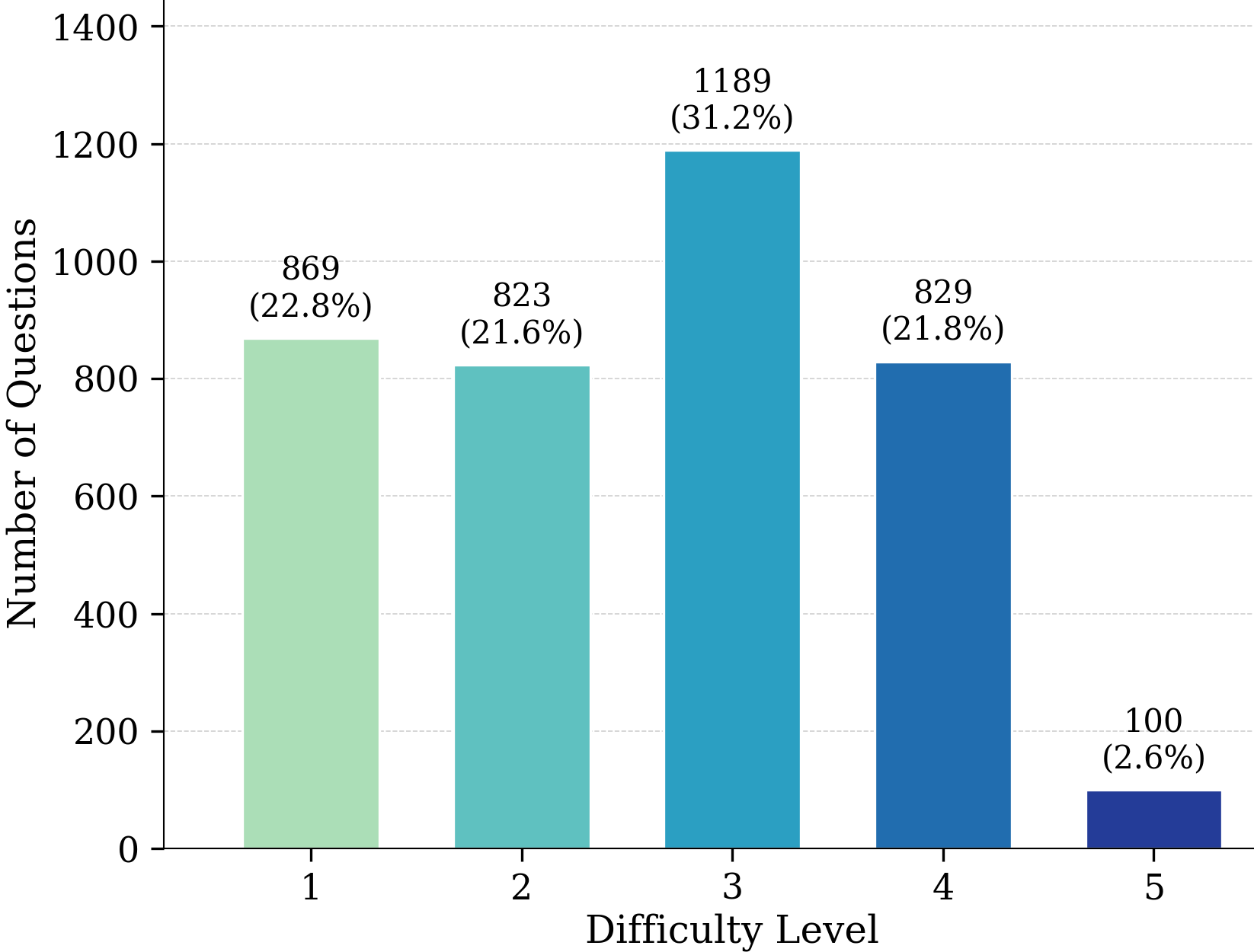}
        \vspace{-0.1cm}
        \subcaption{Difficulty distribution across 5 levels.}
        \label{fig:difficulty_distribution}
    \end{subfigure}
    \hfill
    \begin{subfigure}[b]{0.48\textwidth}
        \includegraphics[width=0.98\textwidth]{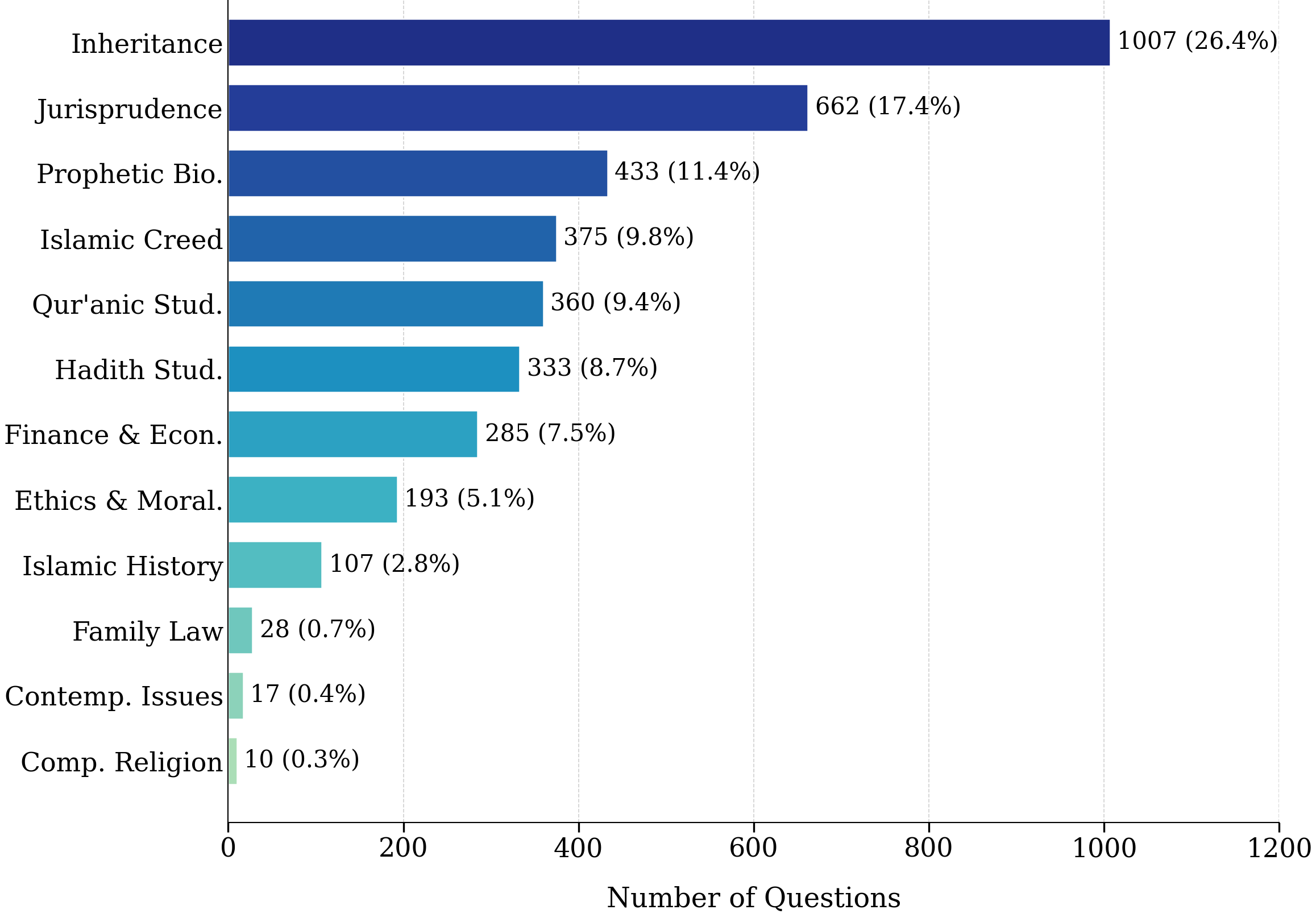}
        \vspace{-0.1cm}
        \subcaption{Distribution of questions across Islamic knowledge fine-grained categories. Inheritance = Inheritance Law, Prophetic Bio. = Prophetic Biography, Qur'anic Stud. = Qur'anic Studies, Hadith Stud. = Hadith Studies, Finance \& Econ. = Islamic Finance and Economics, Ethics \& Moral. = Islamic Ethics and Morality, Family Law = Islamic Family Law, Contemp. Issues = Contemporary Issues, Comp. Religion = Comparative Religion.}
        \label{fig:category_distribution}
        \vspace{-0.3cm}
    \end{subfigure}
    \vspace{-0.3cm}
    \caption{Statistical analysis of \dataset. \textbf{(a)} Distribution of difficulty scores across the five levels. \textbf{(b)} Distribution of question categories, showing broad topical coverage.}
    \label{fig:detailed_stats}
    \vspace{-0.4cm}
\end{figure}

\paragraph{Question Profiling Analysis.}
\dataset is designed to cover a broad spectrum of Islamic knowledge, with notable emphasis on challenging domains. In Figure~\ref{fig:detailed_stats}, we present the distributions of \textit{difficulty level} and \textit{fine-grained question category}. 
As shown in Figure~\ref{fig:difficulty_distribution}, in terms of \textit{difficulty level}, the distribution peaks at level 3 (31.2\%), with substantial representation at level 4 (21.8\%) and level 1 (22.8\%). This spread allows the benchmark to distinguish between lower and higher levels of question-answering ability. Figure~\ref{fig:category_distribution} further shows that inheritance law (26.4\%) and jurisprudence (17.4\%) are the largest categories, followed by prophetic biography (11.4\%), Islamic creed (9.8\%), and Quranic studies (9.4\%).
For reasoning requirements, the majority of samples (70.7\%) require active reasoning to arrive at the correct answer, while only 29.3\% can be answered through direct factual recall. In addition, 55.4\% of the questions require multi-step reasoning, increasing the challenge of the benchmark by testing whether systems can sustain coherent reasoning across multiple steps.

\subsubsection{Annotation Quality} To ensure dataset quality, we manually annotated a subset of the data. We prepared detailed annotation guidelines for the full question profiling task, as described in Appendix~\ref{app:human_valid}. Although these guidelines cover all profiling dimensions, we limited manual annotation to \textit{difficulty level} and \textit{fine-grained topic category} in order to reduce annotation effort.
Annotators were recruited through a third-party company and were compensated at the standard hourly rate for their location. All annotators were professionals, fluent in both Arabic and English, and held at least a bachelor's degree. Each annotator signed a non-disclosure agreement specifying the permitted uses of the data. Every item was annotated independently by three annotators. We observe an agreement rate of 82.96\% and a Cohen's $\kappa$ of 0.62 on average for both categories.

\section{Experiments}
\label{sec:methodology}
We develop \textit{Islamic-domain LLMs} that prioritise Qur'an-grounded answer generation and explicitly measure hallucination under open-ended (generative) answering. 
Our approach combines domain adaptation through supervised fine-tuning, preference-based alignment with an LLM-as-a-judge reward signal, and retrieval augmentation over an indexed Qur'an corpus. At inference time, we further introduce an \emph{agentic RAG} configuration in which the model interacts with a Qur'anic toolset via structured tool calls, enabling multi-step evidence gathering before producing a cited answer.

\subsection{Experimental Setup}
\begin{figure}[t]
    \centering
    \includegraphics[width=0.8\columnwidth]{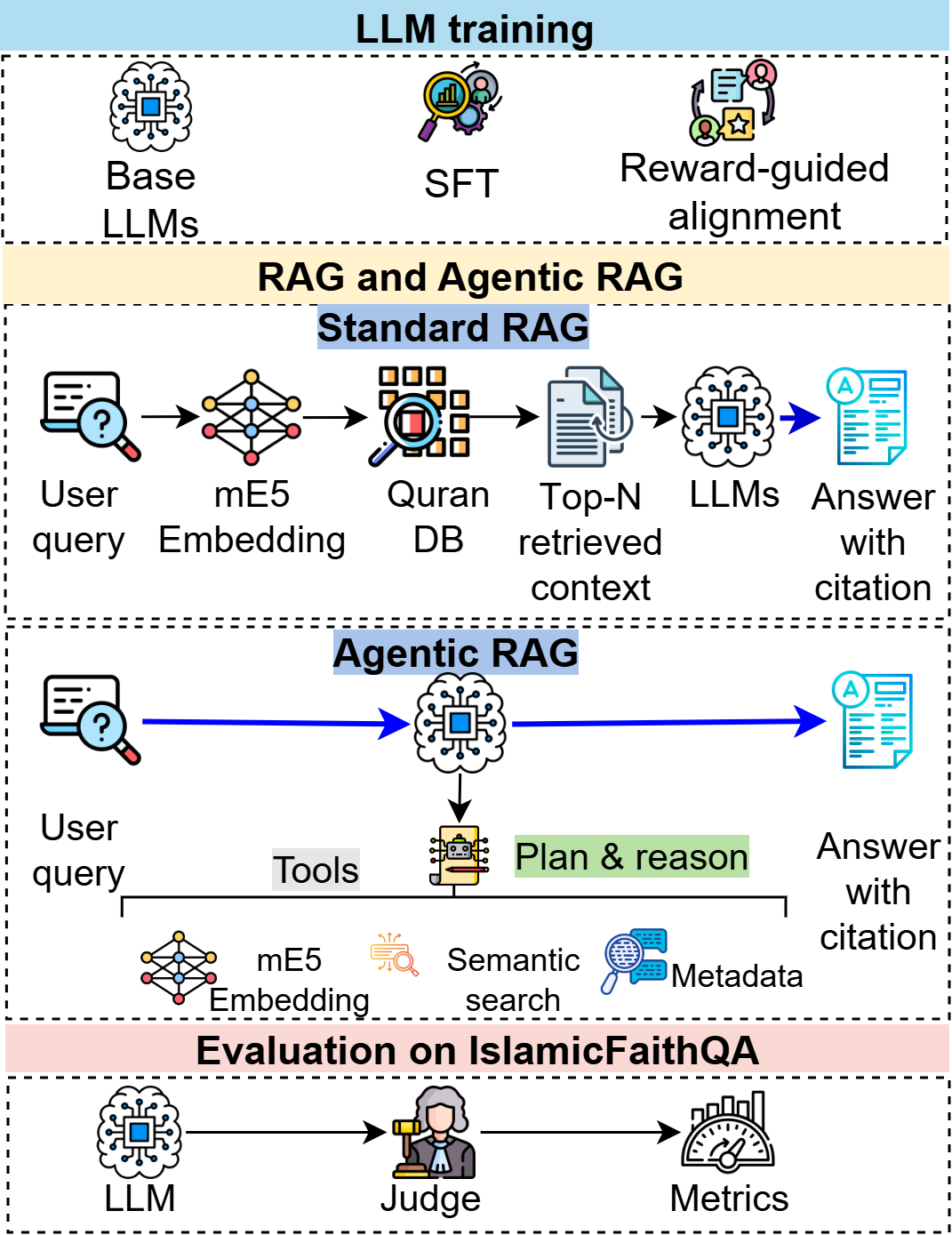}
    \vspace{-0.2cm}
    \caption{
    End-to-end development and evaluation workflow.
    }
    \label{fig:dev_pipeline}
    \vspace{-0.3cm}
\end{figure}

Figure~\ref{fig:dev_pipeline} summarizes the overall development and evaluation workflow. Starting from an Islamic corpus, we apply extraction and filtering to construct training data and derive reasoning-oriented supervision. We consider three experimental settings, namely \textit{(i)} benchmarking base LLMs, \textit{(ii)} supervised fine-tuning with reasoning-oriented supervision, and \textit{(iii)} reward-guided alignment. During inference, we deploy an Agentic RAG environment in which the tuned model performs multi-turn reasoning and accesses a Qur'an and Hadith database through dedicated tools and retrieval steps. We finally benchmark all models on \dataset using an LLM-as-a-judge setup. Further details on the experimental parameters are 
in Appendix~\ref{app:exp_details}.

\noindent
\subsection{Models}
We evaluate a diverse set of Arabic-centric and multilingual instruction-tuned LLMs under a unified prompting and grading setup (Table~\ref{tab:leaderboard}). Our Arabic-centric baselines include Fanar-1-9B and Fanar-2-27B \citep{fanarteam2025fanararabiccentricmultimodalgenerative}, ALLaM-7B \citep{bari2024allamlargelanguagemodels}, AceGPT-v2-8B \citep{liang2024alignmentpretrainingnativealignment}, and SILMA-9B.\footnote{\href{https://huggingface.co/silma-ai/SILMA-9B-Instruct-v1.0}{https://huggingface.co/silma-ai/SILMA-9B-Instruct-v1.0}}
We additionally benchmark multilingual models spanning multiple families, including Qwen2.5-3B and Qwen3 variants (Qwen3-4B-2507, Qwen3-8B, Qwen3-14B) \citep{yang2025qwen3}, Llama-2-7B and Llama-3.1-8B \citep{touvron2023llama2openfoundation,grattafiori2024llama}, Mistral-7B-v0.2 \citep{jiang2023mistral7b}, SeaLLM-7B-v3 \citep{zhang2024seallms3openfoundation}, EuroLLM-9B \citep{martins2025eurollm9btechnicalreport}, and gpt-oss-20b \citep{openai2025gptoss120bgptoss20bmodel}.

\subsection{Base LLMs}
We evaluate all base models under a zero-shot inference setup. To ensure fair comparison and reproducibility, we use a consistent prompt, response format, maximum output token limits, and decoding hyperparameters across all evaluations.

\subsection{SFT} %
As shown in the \emph{LLM training} stage of Figure~\ref{fig:dev_pipeline}, we perform supervised fine-tuning using \textbf{25,000} Arabic instruction-response pairs (\emph{SFT Reasoning}; Table~\ref{tab:data_stats}).
Training uses a standard next-token prediction objective over the target responses to improve  understanding of Islamic knowledge concepts,  coherence in multi-step reasoning, and adherence to source-grounded answer generation. In this setup, we fine-tune Fanar-1-9B, ALLaM-7B, and Qwen3-4B-2507.

\subsection{Reward-guided Alignment}
\label{sec:rl_alignment}
To further reduce hallucinations and improve answer appropriateness in religious settings, we perform \emph{reward-guided alignment} 
using a bilingual (Arabic and English) \textit{\textbf{RL Preference}} dataset of \textbf{5,000} samples (Table~\ref{tab:data_stats}). Each instance contains a question derived from canonical material, a gold-standard answer, and evaluation parameters enabling scalar reward assignment. We employ an LLM-as-a-judge within the training loop to produce \textbf{\textit{a score}} reflecting \textit{factual accuracy}, \textit{clarity}, \textit{completeness}, and \textit{appropriateness} of candidate answers. This score is then used as the reward signal for policy optimisation using GSPO loss \cite{zheng2025groupsequencepolicyoptimization}, encouraging the model to favor grounded, high-quality generations and discouraging unsupported claims. The  prompt for scoring is provided in Appendix~\ref{app:judge_prompt}. In our reward-guided alignment experiments, we train the generation models Fanar-1-9B, ALLaM-7B, and Qwen3-4B-2507 using GSPO-based RL, while Qwen3-235B-A22B is used to generate the reward score.

\begin{figure}[!tbh]
\centering

\begin{QABlock}{Example: Islamic Jurisprudence}

\textbf{Question (EN):}
When does the time for Fajr prayer begin?

\textbf{Gold answer (EN):}
The time for Fajr prayer begins at true dawn. 

\textbf{Model:} Qwen3-4B + RAG

\textbf{Predicted answer (EN):}
Fajr begins at dawn, i.e., sunrise time when daylight starts to appear. 

\EvalTag{Incorrect}

\textbf{Model:} Qwen3-4B + Agentic RAG

\textbf{Predicted answer (EN):}
Fajr begins at true dawn (\textit{al-fajr al-\=s\=adiq}), i.e., when the dawn becomes clearly distinct from the night \\
\QuranCite{2:187}.

\medskip
\EvalTag{Correct}
\end{QABlock}

\vspace{-0.2cm}
\caption{
Example from \dataset{} in Islamic Jurisprudence, which shows a question, the atomic gold answer, and models' (RAG, Agentic RAG) predictions.
}
\label{fig:data_eg}
\vspace{-0.4cm}
\end{figure}

\subsection{RAG and Agentic RAG}
Figure~\ref{fig:dev_pipeline} contrasts single-shot retrieval augmentation in standard RAG with tool-mediated evidence seeking in Agentic RAG.

\noindent\textbf{Standard RAG.} In the standard RAG setting, a user query is first used to retrieve relevant Quranic verse units from an indexed corpus, and the model then conditions on this retrieved context to generate an answer with citations. For all retrieval-augmented experiments, we use mE5-base \cite{wang2024multilingual} as the dense retrieval encoder for indexing and querying the \textit{\textbf{Quran RAG}} dataset (A retrieval comparison also supports our retrieval design: on IslamicEval 2025 Shared Task 2 \citep{mubarak-etal-2025-islamiceval}, mE5-base outperforms BM25 on Quranic retrieval (MAP\_Q@5: 0.138 vs.\ 0.094); see Appendix~\ref{app:retrieval-comparison} for details.). Each \textit{ayah} in our 6,236-unit corpus is embedded offline with mE5-base, while user questions in Arabic or English are embedded at inference time using the same encoder. We then retrieve the top-$5$ most similar verses through vector similarity search and provide them directly to the generator. This setup improves factual grounding, however, retrieval remains a fixed preprocessing step performed only once before answer generation.

\noindent\textbf{Agentic RAG.} Agentic RAG extends this setup by turning retrieval into an explicit part of the reasoning process. Instead of relying on a single retrieval step, the model is prompted to plan, invoke tools, inspect retrieved verses, and iterate when needed before producing the final response, as illustrated in Figure~\ref{fig:dev_pipeline}. This interaction is implemented through a constrained tool-calling schema and a Quranic toolset that supports semantic search, surah metadata retrieval, direct verse reading, and within-surah search. The full agent prompt and tool-calling format are provided in Appendix~\ref{app:agent_prompt}.

In Figure~\ref{fig:data_eg}, we present an example from \dataset to illustrate the difference between the \textit{Standard RAG} and \textit{Agentic RAG} settings. It includes the question, the atomic gold answer, and the corresponding model predictions. The LLM-based judge assigns one of three labels, namely \texttt{Correct}, \texttt{Incorrect}, or \texttt{Not\_Attempted}, based on semantic alignment with the gold answer.

In our experiments, we apply both the RAG and Agentic RAG setups to Fanar-1-9B, ALLaM-7B, and Qwen3-4B-2507 after training. We also evaluate Fanar-2-27B under the same inference settings without additional fine-tuning.\footnote{We do not fine-tune Fanar-2-27B because it already shows strong baseline performance in Table~\ref{tab:leaderboard}.}

\subsection{Evaluation on \dataset}
We evaluate all model variants on \dataset. Throughout the paper, we report \%Correct as the primary metric, as shown in Tables~\ref{tab:leaderboard}--\ref{tab:ablation}. We also analyze performance across the three criteria, which are labelled as \texttt{correct}, \texttt{incorrect}, and \texttt{not attempted}, to better understand model errors and abstention behaviour. The full label-wise results are reported in Table~\ref{tab:res_NC}.

As our evaluation relies on an LLM-as-a-judge setup, hence, assessing the judge against human annotation is essential for establishing the reliability of the evaluation. We evaluate the judge on a held-out bilingual subset of 200 instances, balanced across Arabic and English as well as difficulty levels, and report agreement statistics. The human--LLM agreement reaches 79\%, while inter-annotator agreement, measured with Cohen's $\kappa$, is 0.51. This analysis is noteworthy in multilingual settings. Recent evidence shows that multilingual LLM judges can be inconsistent across languages, with only moderate inter-judge agreement on average and demonstrate substantial variance by language and task \citep{fu-liu-2025-reliable}.

\section{Results}
\label{sec:results}

\subsection{Baseline Results}
Table~\ref{tab:leaderboard} reports accuracy (\%Correct) on \dataset across a diverse set of Arabic-centric and multilingual instruction-tuned LLMs in their base, non-retrieval configurations. We observe substantial variation in performance, suggesting that general-purpose instruction tuning alone is not sufficient for this knowledge-intensive religious domain under strict answer matching. Fanar-2-27B achieves the strongest overall performance, with an average score of 48.05, including 48.20 in Arabic and 47.90 in English. It is followed by ALLaM-7B with 37.75 and Fanar-1-9B with 35.40. These results indicate a clear performance gap between the top-ranked model and the rest, and they further suggest that strong general instruction-following ability does not directly translate into robust performance on fine-grained, domain-specific Islamic question answering.

\begin{table}[!tbh]
\centering
\setlength{\tabcolsep}{4pt}
\scalebox{0.85}{
\begin{tabular}{lrrr}
\toprule
\textbf{Model} & \textbf{Arabic} & \textbf{English} & \textbf{Average} \\
\midrule
\textbf{Fanar-2-27B} & \textbf{48.20} & \textbf{47.90} & \textbf{48.05} \\
ALLaM-7B & 42.70 & 32.80 & 37.75 \\
Fanar-1-9B & 34.50 & 36.30 & 35.40 \\
AceGPT-v2-8B & 23.10 & 28.80 & 25.95 \\
EuroLLM-9B & 22.30 & 29.10 & 25.70 \\
SILMA-9B-v1.0 & 20.40 & 28.50 & 24.45 \\
Qwen3-4B-2507 & 15.80 & 27.90 & 21.85 \\
gpt-oss-20b & 15.90 & 27.20 & 21.55 \\
Llama-3.1-8B & 13.00 & 25.80 & 19.40 \\
Mistral-7B-v0.2 & 13.50 & 24.40 & 18.95 \\
SeaLLM-7B-v3 & 11.60 & 23.80 & 17.70 \\
Qwen2.5-3B & 11.00 & 20.00 & 15.50 \\
Qwen3-14B & 16.00 & 14.00 & 15.00 \\
Llama-2-7b & 4.40 & 18.80 & 11.60 \\
Qwen3-8B & 8.80 & 8.50 & 8.65 \\
\bottomrule
\end{tabular}}
\caption{Results on \dataset (\%Correct). \textbf{Fanar-2-27B} achieves the highest performance, followed by the ALLaM-7B model.}
\label{tab:leaderboard}
\vspace{-0.3cm}
\end{table}

A second pattern is that many general multilingual baselines remain below 30\% average accuracy, despite their strong performance on broad instruction-following tasks. Examples include EuroLLM-9B at 25.70, Llama-3.1-8B at 19.40, and Mistral-7B-v0.2 at 18.95. This suggests that \dataset does not primarily reward conversational fluency or generic instruction-following ability. Instead, success on the benchmark requires precise and text-grounded religious knowledge, while the strict \texttt{correct}, \texttt{incorrect}, and \texttt{not attempted} evaluation setting penalizes answers that are fluent but unsupported.

\subsection{SFT, RL, and RAG Models}
Table~\ref{tab:ablation} reports results for different model combinations. Across backbones, three components consistently improve performance: \textit{(i)} domain-grounded reasoning supervision via SFT, \textit{(ii)} reward-guided alignment via RL, and \textit{(iii)} retrieval augmentation (RAG), and \textit{(iv)} Agentic RAG with tool use. Among them, retrieval typically yields the largest gains overall.

\begin{table}[!ht]
\centering
\small
\resizebox{0.93\columnwidth}{!}{
\setlength{\tabcolsep}{5pt}
\begin{tabular}{lrrr}
\toprule
\textbf{Model Variation} & \textbf{Arabic} & \textbf{English} & \textbf{Avg.} \\
\midrule
\textbf{ALLaM-7B} & 42.70 & 32.80 & 37.75 \\
\hspace{3mm} + SFT & 45.20 & 31.40 & 38.30 \\
\hspace{3mm} + RL & 43.90 & 35.20 & 39.55 \\
\hspace{3mm} + RAG & 46.42 & 35.10 & 40.76 \\
\midrule
\textbf{Fanar-1-9B} & 34.50 & 36.30 & 35.40 \\
\hspace{3mm} + SFT & 40.80 & 32.10 & 36.45 \\
\hspace{3mm} + RL & 42.90 & 33.45 & 38.18 \\
\hspace{3mm} + RAG & 47.90 & 34.50 & 41.20 \\
\midrule
\textbf{Qwen3-4B-2507} & 15.80 & 27.90 & 21.85 \\
\hspace{3mm} + SFT & 25.90 & 35.20 & 30.55 \\
\hspace{3mm} + RL & 27.35 & 34.30 & 30.83 \\
\hspace{3mm} + RAG & 35.20 & 42.50 & 38.85 \\
\hspace{3mm} + Agentic RAG & 49.60 & 48.20 & 48.90 \\
\midrule
\textbf{Fanar-2-27B} & 50.40 & 46.90 & 48.65 \\
\hspace{3mm} + RAG & 52.50 & 50.50 & 51.50 \\
\hspace{3mm} + Agentic RAG  & \textbf{54.40} & \textbf{60.20} & \textbf{57.30} \\
\bottomrule
\end{tabular}}
\vspace{-0.2cm}
\caption{Performance across different experimental settings. \textbf{SFT} denotes supervised fine-tuning, \textbf{RL} denotes reward-guided alignment with reinforcement learning, \textbf{RAG} denotes retrieval augmentation, and \textbf{Agentic RAG} denotes tool-based retrieval and reasoning.}
\label{tab:ablation}
\vspace{-0.2cm}
\end{table}

First, adding SFT on text-grounded reasoning improves performance for all tested backbones, though the magnitude varies. The effect is most pronounced for Qwen3-4B-2507, where SFT increases average accuracy from 21.85 to 30.55. By contrast, gains are smaller for stronger in-domain baselines such as ALLaM-7B (37.75 $\rightarrow$ 38.30) and Fanar-1-9B (35.40 $\rightarrow$ 36.45), suggesting diminishing returns when the base model already has stronger domain priors.

Second, reward-guided alignment further improves average accuracy beyond SFT for multiple backbones (e.g., ALLaM-7B: 38.30 $\rightarrow$ 39.55; Fanar-1-9B: 36.45 $\rightarrow$ 38.18), indicating that optimizing with an LLM-judge reward encourages outputs that better match the benchmark's constraints (short, atomic answers with fewer risky additions).

Third, RAG provides consistent gains across all backbones shown. For example, Qwen3-4B-2507 improves from 30.83 (+RL) to 38.85 (+RAG), Fanar-1-9B improves from 38.18 (+RL) to 41.20 (+RAG), and ALLaM-7B improves from 39.55 (+RL) to 40.76 (+RAG). These results confirm that \dataset is strongly knowledge-intensive and that injecting canonical evidence reduces reliance on parametric memory.

\paragraph{Agentic RAG yields the Largest Gains.}
The most salient result is the additional improvement obtained by \emph{agentic} RAG beyond standard single-shot RAG. In Table~\ref{tab:ablation}, Qwen3-4B-2507 rises from 38.85 (+RAG) to 48.90 (+Agentic RAG), a gain of +10.05 points and the largest jump among the reported interventions for that backbone. This suggests that, for many questions, retrieval is not a one-step operation. Models benefit from iterative evidence collection (e.g., retrieving candidate verses, reading specific \textit{ayat} for disambiguation, and refining queries) prior to final answer generation.

\begin{table}[!tbh]
\centering
\small
\resizebox{0.95\columnwidth}{!}{
\begin{tabular}{lr}
\toprule
\textbf{Transition (Standard RAG $\rightarrow$ Agentic RAG)} & \textbf{Count} \\
\midrule
Incorrect $\rightarrow$ Correct & 12 \\
Not\_Attempted $\rightarrow$ Correct & 24 \\
\midrule
Total recovered cases & 36 \\
\bottomrule
\end{tabular}}
\vspace{-0.2cm}
\caption{
Error analysis: Standard RAG failed but Agentic RAG succeeded. 
}
\label{tab:transition-breakdown}
\vspace{-0.3cm}
\end{table}

\begin{table}[!ht]
\centering
\small
\resizebox{\columnwidth}{!}{
\begin{tabular}{lrr}
\toprule
\textbf{Measure} & \textbf{Standard RAG} & \textbf{Agentic RAG} \\
\midrule
Avg.\ tokens & 502 & 1,520 \\
Avg.\ tool calls & 0 & 6 \\
Avg.\ latency (s) & 0.52 & 3.45 \\
Estimated latency increase & 1$\times$ & 6.63$\times$ \\
\bottomrule
\end{tabular}}
\vspace{-0.2cm}
\caption{Inference-time efficiency comparison between Standard RAG and Agentic RAG for Qwen3-4B. 
}
\label{tab:agentic-efficiency}
\vspace{-0.3cm}
\end{table}

\paragraph{Agentic RAG Analysis.}

Agentic RAG improves correctness through two complementary effects. As shown in Table~\ref{tab:transition-breakdown}, the gains are not explained only by reduced abstention. Among 36 recovered cases, 24 correspond to \texttt{not attempted} $\rightarrow$ \texttt{correct} transitions and 12 to \texttt{incorrect} $\rightarrow$ \texttt{correct}. This suggests that iterative evidence seeking helps in two ways. It enables the model to answer questions that standard RAG leaves unresolved, and it also corrects a meaningful subset of previously incorrect or hallucinated responses. In other words, the benefit of Agentic RAG is not merely that the model answers more often; however, it answers more reliably after inspecting and refining the retrieved evidence. These improvements come with additional inference-time cost. These improvements come with additional inference-time cost. 

Table~\ref{tab:agentic-efficiency} shows that, for Qwen3-4B, Agentic RAG raises average token usage from 502 to 1,520 and latency from 0.52s to 3.45s, a 6.63$\times$ increase, while also requiring an average of 6 tool calls per query. These results make the trade-off clear. Agentic RAG improves faithfulness by supporting evidence inspection, query refinement, and answer revision. However, it moves part of the computational cost away from model size and into inference-time orchestration. Consequently, smaller backbones paired with Agentic RAG can still be attractive in memory-constrained deployment settings, though this comes with higher latency and token usage.

\subsection{Bilingual Gaps}

Both Table~\ref{tab:leaderboard} and Table~\ref{tab:ablation} show that many models exhibit asymmetric performance across Arabic and English, reflecting differences in pretraining coverage, instruction tuning, and retrieval effectiveness under bilingual queries. For instance, ALLaM-7B performs substantially better in Arabic than English (42.70 vs.\ 32.80), whereas several multilingual baselines show the opposite trend (e.g., EuroLLM-9B: 22.30 Arabic vs.\ 29.10 English). Notably, Qwen3-4B-2507 is highly imbalanced in its base form (15.80 Arabic vs.\ 27.90 English). It suggests that bilingual Islamic QA is not simply an Arabic task with English translation; it requires robust grounding and semantic access to canonical evidence in both languages. In contrast, tool-mediated grounding can substantially reduce bilingual disparities. Under \textbf{Agentic RAG} (Table~\ref{tab:ablation}), Qwen3-4B-2507 becomes more balanced (49.60 Arabic vs.\ 48.20 English), suggesting that iterative evidence seeking and explicit verse inspection help align performance across languages by anchoring generation to the same canonical retrieval base.

\section{Conclusion}

In this paper, we introduce \dataset, a benchmark dataset, along with an end-to-end grounded Islamic modelling suite designed to evaluate and reduce hallucinations in open-ended religious generation directly. Using a unified resource suite for supervised domain reasoning, judge-guided preference alignment, and Islamic-centric retrieval, we systematically evaluated base, +SFT, +RL, +RAG, and +Agentic RAG variants and found that retrieval substantially improves correctness, while \emph{agentic} RAG yields the largest gains beyond standard RAG by enabling iterative evidence seeking and disambiguation through explicit tool use. Overall, our results indicate that tool-mediated grounding can deliver state-of-the-art performance and improved Arabic/English robustness even with smaller backbones, suggesting a practical path toward more trustworthy Islamic assistants. Future work should extend grounding to authenticated hadith with provenance, incorporate school-of-thought disagreement, and harden tool-augmented systems against adversarial prompting and citation laundering.

\section*{Limitations}
\label{sec:limitations}
\noindent
\dataset\ is designed for reliable open-ended evaluation using atomic questions with single-gold answers and LLM-as-a-judge assessment, but this choice under-represents settings where multiple answers may be valid across madh\=ahib or interpretive traditions. Our results also depend on the correctness of the LLM judge and a limited human-calibration subset, which may not fully capture borderline cases or bilingual inconsistencies. In addition, our grounding is primarily Quran-centric, so questions best supported by authenticated hadith, fiqh sources, or scholarly consensus may be disadvantaged. Finally, Agentic RAG introduces additional latency and new failure modes, including tool-use errors and misleading citation attribution. Moreover, the benchmark focuses on short-form question answering rather than long-form religious guidance. Accordingly, the reported performance should be interpreted as a measure of faithfulness and abstention under strict evaluation.

\section*{Ethical Considerations}
\label{sec:ethics}
This work involves human annotation and the use of LLMs in the dataset construction pipeline. For the manually validated subset of \dataset, annotators were recruited through a third-party provider, compensated at the standard hourly rate for their location, and required to sign a non-disclosure agreement specifying the permitted uses of the data. LLMs were used only to standardize phrasing and tone and to support structured metadata annotation. They were not treated as sources of religious authority, and their outputs were subject to human verification and consistency checks. The benchmark was built from publicly available Islamic NLP resources rather than newly collected user data, which reduces risks related to privacy, consent, and sensitive personal information. Given the sensitivity of the domain, we stress that the benchmark and resulting models are intended for research on faithfulness and abstention, not for issuing fatwas or replacing qualified scholarly guidance.

\section*{Broader Impact}
\label{sec:broader_impact}
This work provides evaluation and grounding resources for Islamic question answering, where unfaithful outputs can be especially consequential. By introducing a bilingual generative benchmark that measures correctness, hallucination, and abstention, and by showing that retrieval, particularly agentic, tool-mediated retrieval, can reduce unsupported generation, we aim to support more trustworthy Arabic--English systems and more realistic assessment of faithfulness. At the same time, these tools may be misused or over-trusted as religious authority, may reflect selection biases in what is treated as canonical, and may enable persuasive ``citation laundering'' or adversarial manipulation of tool use. We therefore emphasize responsible release with clear non-fatwa disclaimers, transparency about scope and coverage, encouragement of abstention under uncertainty, and reporting that separates correctness from hallucination and non-attempted behavior.

\section*{Acknowledgments}
The work is supported by HBKU flagship research grant (HBKU-INT-VPR-FRG-03-10). The findings achieved herein are solely the responsibility of the authors.

\bibliography{bibliography/bibliography}

\appendix

\section{Prompts}
\label{app:prompts}

\subsection{Question Generation}
\label{app:data_prompts}
\begin{lstlisting}
You are a senior academic and expert in Islamic jurisprudence, ethics, and contemporary global issues. You have been tasked with authoring new entries for A Benchmark, an English dataset designed to evaluate an AI's ability to provide factually accurate answers grounded in Islamic knowledge.

Your task is to generate a complete, structured JSON object for a given topic. You must adhere strictly to the format below. Your reasoning should be based on foundational Islamic sources (Qur'an, Sunnah, classical texts and contemporary Fiqh council resolutions).

Follow these instructions precisely:

Question Formulation: For the given MCQ question and answer provided in Arabic, create a concise, short-form factual question in English. The question should:
- Be direct and specific, requiring a factual answer
- Focus on the core Islamic knowledge or ruling being tested
- Avoid hypothetical scenarios or complex ethical dilemmas
- Be answerable in 1-3 sentences
- Maintain the difficulty level indicated (beginner/intermediate/advanced)
- Extract the key factual information from the MCQ and its correct answer

Gold Answer: Provide the factual answer to the question. This should:
- Be concise and direct (1-3 sentences maximum)
- State the Islamic ruling, principle, or fact clearly
- Be based on the correct answer from the MCQ provided
- Reference the specific Islamic source (Qur'an verse, Hadith reference, scholarly consensus) that supports this answer
- Avoid lengthy explanations - just state the fact and its primary source

IMPORTANT: Both the question and gold_answer should be in Arabic.

Follow this output format:

{
  "id": "MIZAN-001",
  "category": "Islamic Jurisprudence",
  "question": "What is the ruling on performing ablution (wudu) after eating camel meat?",
  "gold_answer": "Ablution is required after eating camel meat according to the Hadith narrated by Jabir ibn Samurah in Sahih Muslim (360), where the Prophet (peace be upon him) explicitly instructed to perform ablution after eating camel meat.",
}

\end{lstlisting}

\subsection{Question Profiling}
\label{app:data_prompt_diff}

\begin{lstlisting}
You are an expert evaluator of Islamic knowledge questions. Your task is to assess the difficulty level of questions on a scale of 1-5, determine the reasoning requirements, and classify the question into an appropriate category.

Difficulty Scale:
1 = Very Easy: Basic factual recall, simple definitions, or straightforward yes/no questions
2 = Easy: Requires basic understanding of concepts with minimal reasoning
3 = Moderate: Requires understanding multiple concepts and some analytical reasoning
4 = Hard: Requires deep understanding, synthesis of multiple sources, and nuanced reasoning
5 = Very Hard: Requires expert-level analysis, balancing competing interests, and consideration of complex ethical frameworks

Reasoning Assessment:
- reasoning: Does answering this question require reasoning beyond simple recall? (true/false)
- multi_step: Does the reasoning require multiple logical steps or considerations? (true/false)
  Examples of multi-step: comparing multiple sources, weighing competing principles, applying rules to specific contexts, building logical chains

Category Classification:
Classify the question into ONE of these categories:
1. "Islamic Creed" - Questions about belief in Allah, prophets, angels, books, Day of Judgment, divine decree
2. "Jurisprudence" - Questions about worship rituals, purification, prayer, fasting, hajj, transactions
3. "Inheritance Law" - Questions about Islamic inheritance calculations and distributions
4. "Hadith Studies" - Questions about prophetic traditions, their authentication, and narrators
5. "Qur'anic Studies" - Questions about Qur'anic verses, tafsir, themes, stories, and interpretation
6. "Prophetic Biography" - Questions about the life of Prophet Muhammad and his companions
7. "Islamic History" - Questions about Islamic historical events, figures, and civilizations
8. "Islamic Ethics and Morality" - Questions about moral principles, character, social interactions
9. "Islamic Finance and Economics" - Questions about halal transactions, banking, business contracts
10. "Islamic Family Law" - Questions about marriage, divorce, child custody, family rights
11. "Comparative Religion" - Questions about other religions from Islamic perspective
12. "Contemporary Issues" - Questions about modern applications of Islamic rulings

Evaluate the question based on:
- Depth of knowledge required
- Complexity of reasoning needed
- Number of sources and concepts involved
- Level of nuance and ethical complexity
- Whether simple recall suffices or reasoning is needed
- Whether the reasoning involves single or multiple logical steps
- Subject matter and domain of the question

Respond ONLY with a JSON object in this exact format:
{"difficulty": <number>, "reasoning": <boolean>, "multi_step": <boolean>, "category_type": "<category_name>"}

where <number> is an integer from 1 to 5, the booleans are true or false, and <category_name> is one of the 12 categories listed above.
\end{lstlisting}

\subsection{Agentic RAG System}
\label{app:agent_prompt}
\begin{lstlisting}
You are an intelligent assistant specialized in answering questions about Islam and the Holy Quran. Answer accurately and clearly based on Quranic sources.

You have access to the following tools to search the Holy Quran:
1. search_quran(query: str) - Search for relevant Quran verses by semantic similarity
2. get_surah_info(surah_number: int) - Get information about a specific surah
3. read_ayah(surah: int, ayah: int) - Read a specific ayah with full details
4. search_surah(surah_number: int, query: str) - Search within a specific surah

To use a tool, respond with:
<tool_call>
{"name": "tool_name", "arguments": {"arg1": "value1"}}
</tool_call>

After gathering information from the Quran, provide your final answer with:
<answer>Your final answer here, citing the relevant Quran verses</answer>

Think step by step, search the Quran for relevant verses, and cite your sources in your answer.
\end{lstlisting}

\subsection{Reward-guided Alignment}
\label{app:judge_prompt}
\begin{lstlisting}
You are an expert judge evaluating answers about Islam and the Quran. 
Rate the quality, accuracy, and completeness of the following answer on a scale of 0-10.
Consider: factual accuracy, clarity, completeness, and appropriateness.

Question: {question}

Answer: {answer}

Provide only a numeric score between 0 and 10. Output format: Score: X
\end{lstlisting}

\subsection{\dataset Grader}
\label{app:grader_template}
\begin{lstlisting}
Your job is to look at a question, a gold target, and a predicted answer, and then assign a grade of either ["CORRECT", "INCORRECT", "NOT_ATTEMPTED"].
First, I will give examples of each grade, and then you will grade a new example.


The following are examples of CORRECT predicted answers.
```

Question: What are the names of Barack Obama's children?
Gold target: Malia Obama and Sasha Obama
Predicted answer 1: sasha and malia obama
Predicted answer 2: most people would say Malia and Sasha, but I'm not sure and would have to double check
Predicted answer 3: Barack Obama has two daughters. Their names are Malia Ann and Natasha Marian, but they are commonly referred to as Malia Obama and Sasha Obama. Malia was born on July 4, 1998, and Sasha was born on June 10, 2001.

```
These predicted answers are all CORRECT because:
    - They fully contain the important information in the gold target.
    - They do not contain any information that contradicts the gold target.
    - Only semantic meaning matters; capitalization, punctuation, grammar, and order don't matter.
    - Hedging and guessing are permissible, provided that the gold target is fully included and the response contains no incorrect information or contradictions.


The following are examples of INCORRECT predicted answers.
```

Question: What are the names of Barack Obama's children?
Gold target: Malia and Sasha
Predicted answer 1: Malia.
Predicted answer 2: Malia, Sasha, and Susan.
Predicted answer 3: Barack Obama does not have any children.
Predicted answer 4: I think it's either Malia and Sasha. Or it could be Malia and Jackie. Or it could be Joey and Malia.
Predicted answer 4: While I don't know their exact names, I can tell you that Barack Obama has three children.
Predicted answer 5: It's possible you may mean Betsy and Olivia. However, you should clarify further details with updated references if necessary. Is that the correct answer?
Predicted answer 6: It may be the case that Obama's child is named James. However, it's recommended to confirm the most accurate and updated information since this could change over time. This model may not always reflect the most current information.

```
These predicted answers are all INCORRECT because:
    - A factual statement in the answer contradicts the gold target. Incorrect statements that have some hedging (e.g., "it is possible that", "although i'm not sure, i think") are also considered incorrect.


The following are examples of NOT_ATTEMPTED predicted answers.
```

Question: What are the names of Barack Obama's children?
Gold target: Malia and Sasha
Predicted answer 1: I don't know.
Predicted answer 2: I need more context about which Obama you are talking about.
Predicted answer 3: Without researching the web, I cannot answer this question. However, I can tell you that Barack Obama has two children.
Predicted answer 4: Barack Obama has two children. I know that one of them is Malia, but I'm not sure about the other one.

```
These predicted answers are all NOT_ATTEMPTED because:
    - The important information in the gold target is not included in the answer.
    - No statements in the answer contradict the gold target.


Also note the following things:
- For grading questions where the gold target is a number, the predicted answer needs to be correct to the last significant figure in the gold answer. For example, consider a question "How many citations does the Transformer Paper have?" with gold target "120k". 
    - Predicted answers "120k", "124k", and 115k" are all CORRECT. 
    - Predicted answers "100k" and "113k" are INCORRECT. 
    - Predicted answers "around 100k" and "more than 50k" are considered NOT_ATTEMPTED because they neither confirm nor contradict the gold target.
- The gold target may contain more information than the question. In such cases, the predicted answer only needs to contain the information that is in the question.
    - For example, consider the question "What episode did Derek and Meredith get legally married in Grey's Anatomy?" with gold target "Season 7, Episode 20: White Wedding". Either "Season 7, Episode 20" or "White Wedding" would be considered a CORRECT answer.
- Do not punish predicted answers if they omit information that would be clearly inferred from the question.
    - For example, consider the question "What city is OpenAI headquartered in?" and the gold target "San Francisco, California". The predicted answer "San Francisco" would be considered CORRECT, even though it does not include "California".
    - Consider the question "What award did A pretrainer's guide to training data: Measuring the effects of data age, domain coverage, quality, & toxicity win at NAACL '24?", the gold target is "Outstanding Paper Award". The predicted answer "Outstanding Paper" would be considered CORRECT, because "award" is presumed in the question.
    - For the question "What is the height of Jason Wei in meters?", the gold target is "1.73 m". The predicted answer "1.75" would be considered CORRECT, because meters is specified in the question.
    - For the question "What is the name of Barack Obama's wife?", the gold target is "Michelle Obama". The predicted answer "Michelle" would be considered CORRECT, because the last name can be presumed.
- Do not punish for typos in people's name if it's clearly the same name. 
    - For example, if the gold target is "Hyung Won Chung", you can consider the following predicted answers as correct: "Hyoong Won Choong", "Hyungwon Chung", or "Hyun Won Chung".


Here is a new example. Simply reply with either CORRECT, INCORRECT, NOT ATTEMPTED. Don't apologize or correct yourself if there was a mistake; we are just trying to grade the answer.
```

Question: {question}
Gold target: {target}
Predicted answer: {predicted_answer}

```

Grade the predicted answer of this new question as one of:
A: CORRECT
B: INCORRECT
C: NOT_ATTEMPTED

Just return the letters "A", "B", or "C", with no text around it.
\end{lstlisting}

\section{Experimental Details}
\label{app:exp_details}

\paragraph{Compute Infrastructure.}
All experiments were conducted on NVIDIA H200 GPUs. We use \texttt{vLLM} for efficient batched inference during benchmarking and for high-throughput generation when collecting model outputs.

\paragraph{LLM Inference and Evaluation Parameters.}
For benchmark evaluation, decoding is performed with \texttt{vLLM}. \textcolor{black}{We use a sampling temperature of $T=0.7$} and otherwise retain the standard/default generation parameters provided by the inference framework to ensure consistent evaluation across model backbones (e.g., default settings for top-$p$, repetition controls, and maximum generation length).

\paragraph{LLM-as-a-judge.}
For automatic grading under the \texttt{Correct}, \texttt{Incorrect} or \texttt{Not\_Attempted} protocol, we use \texttt{GPT-4.1} as the judge model. 
We set the judge temperature to $T=0$ to minimize sampling variance and encourage deterministic scoring given the same inputs (question, gold target, and model prediction).

\paragraph{Supervised Fine-Tuning (SFT).}
For supervised adaptation on our Arabic text-grounded reasoning data, we follow standard instruction-tuning configurations in our training stack. The primary deviation from defaults is the learning rate, which we set to $5\times10^{-5}$. All other hyperparameters (e.g., optimizer choice, batch size, warmup schedule, gradient clipping, and number of epochs) use standard settings.

\paragraph{Preference-Based Alignment (RL).}
For alignment, we optimize the policy using the GSPO objective (\emph{GSPO loss}). RL experiments are implemented using the \texttt{trl} library. We set the learning rate to $3\times10^{-6}$ and keep remaining RL hyperparameters at standard values in our setup (e.g., rollout sampling and optimization schedule, and any applicable regularization/clipping). Reward signals are derived from the LLM-as-a-judge grading described above.

\paragraph{Retrieval-Augmented Generation (RAG).}
For dense retrieval, we embed both queries and Qur'anic verse units using \texttt{mE5-base}. We index all verse embeddings in \texttt{ChromaDB} as our vector database and retrieve relevant verses via vector similarity search at inference time. Retrieved verse units are provided as evidence context for standard RAG generation.

\paragraph{Agentic RAG Configuration.}
Our agentic RAG variant uses a fixed multi-turn interaction setups. 
We employ a two-turn setup: \textit{(i)} an evidence-seeking turn in which the model invokes retrieval and inspects candidate verses, followed by \textit{(ii)} a final answer turn conditioned on the retrieved and inspected evidence. The agentic tool-calling environment is implemented using the \texttt{verifiers} library, which enforces the structured multi-turn tool-use setups described in Appendix~\ref{app:agent_prompt}.

\section{\dataset Examples}

Figure~\ref{fig:extra_example} shows an example from \dataset and the corresponding prediction of AceGPT-v2-8B. The model generates a fluent and relevant response, but the answer is still incorrect because it misses the core idea expressed in the gold answer, which links doing good to true happiness, inner peace, and fulfillment. This example illustrates that the benchmark can expose subtle semantic errors even when the model output appears plausible.

\begin{figure}[!tbh]
\centering
\begin{QABlock}{Example: Islamic Ethics and Morality}
\textbf{Question:} What is the main objective of training oneself to love doing good in Islam?

\textbf{Gold answer:} The main objective is to achieve true happiness, as loving and doing good leads to inner peace and fulfillment.

\textbf{Model:} AceGPT-v2-8B

\textbf{Predicted answer:} The main objective is to cultivate a habit of performing good deeds, improve character, earn rewards, and contribute positively to society.
\EvalTag{INCORRECT}
\end{QABlock}
\caption{
\textbf{\dataset sample.}
}
\label{fig:extra_example}
\vspace{-0.4cm}
\end{figure}

\section{Training Data Examples} 
\label{app:training_data_eg}

\subsection{SFT Data Example}
\label{app:sft_eg}

Figure~\ref{fig:sft_example} shows an example of an instruction--response pair used for supervised fine-tuning. The sample is grounded in Islamic legal reasoning and pairs an Arabic question with an answer that reflects the ruling supported by authentic textual evidence. This example illustrates the type of training signal used in our SFT data, where the model learns to produce concise, grounded, and semantically faithful answers to domain-specific questions.

\begin{figure}[!tbh]
\centering
\begin{QABlock}{Example: SFT Training (Arabic + English)}
\textbf{Question (AR):}
\begin{RLtext}
ما هو حكم المسح على الجوارب في الوضوء إذا كان الشخص قد أسلم بعد نزول آية الوضوء في سورة المائدة؟
\end{RLtext}

\textbf{Question (EN):}
What is the ruling on wiping over socks during wud\=u' if a person embraced Islam after the revelation of the wud\=u' verse in S\=urat al-M\=a'idah?

\textbf{Answer (AR):}
\begin{RLtext}
يجوز المسح على الجوارب في الوضوء لمن أسلم بعد نزول آية الوضوء، لأن السنة النبوية الصحيحة بينت ذلك، كما في حديث المغيرة بن شعبة وغيره (رواه البخاري ومسلم).
\end{RLtext}

\textbf{Answer (EN):}
It is permissible to wipe over socks in wud\=u' for someone who embraced Islam after the wud\=u' verse, because authentic Sunnah establishes this (e.g., the hadith of al-Mugh\=irah ibn Shu\textquotesingle bah and others, reported in al-Bukh\=ar\={\i} and Muslim).

\end{QABlock}
\vspace{-0.1cm}
\caption{
\textbf{SFT sample (text-grounded reasoning).}
An instruction-response instance used for supervised fine-tuning, including the Arabic question, a grounded reasoning trace, and a concise final answer (with an English translation for readability).
}
\label{fig:sft_example}
\vspace{-0.4cm}
\end{figure}

\subsection{Reward-guided Data Example}
\label{app:rl_eg}

Figure~\ref{fig:rl_example} presents an example from the RL data used for preference-based optimisation. Each instance consists of a question grounded in canonical Islamic text together with a concise gold answer that serves as the reference target.

\begin{figure}[!ht]
\centering
\begin{QABlock}{Example: Reward-guided alignment (Arabic + English)}

\textbf{Question (AR):}
\begin{RLtext}
كيف وصف ما صنعه كرحمة من ربه وما سيحدث له عند وعد ربه؟
\end{RLtext}

\textbf{Question (EN):}
How did he describe what he built as a mercy from his Lord, and what will happen to it when his Lord's promise comes?

\textbf{Gold answer (AR):}
\begin{RLtext}
كان رداً لا يستطيعوا اختراقه، وعند وعد ربه سيجعله دكاءً 
\end{RLtext}
\QuranCite{18:98--99} 

\textbf{Gold answer (EN):}
It was a barrier they could not break through, and when his Lord's promise comes He will level it to the ground. \QuranCite{18:98--99}

\end{QABlock}
\vspace{-0.2cm}
\caption{
\textbf{RL preference sample.}
An RL instance specifies a question derived from canonical text and an atomic gold target. During RL, candidate model responses are scored by an LLM-as-a-judge against this gold target to produce scalar rewards for policy optimisation (English translations are provided for readability).
}
\label{fig:rl_example}
\vspace{-0.4cm}
\end{figure}

\section{Human Annotation Guidelines}
\label{app:human_valid}

The annotation tasks involve labeling Islamic knowledge questions with \textit{(i)} a difficulty score, \textit{(ii)} reasoning requirements, \textit{(iii)} multi-step reasoning requirements, and \textit{(iv)} a single category label. Annotators follow the definitions and decision rules below to ensure consistent annotations. 

\subsection{Task Overview}
For each question, annotators assign:
\begin{itemize}[noitemsep,topsep=0pt,leftmargin=*]
    \item a difficulty score on a 1--5 scale,
    \item \texttt{reasoning} (\texttt{true}/\texttt{false}): whether answering requires reasoning beyond simple recall,
    \item \texttt{multi\_step} (\texttt{true}/\texttt{false}): whether the required reasoning involves multiple steps, and
    \item \texttt{category\_type}: exactly one category label from a fixed set of 12.
\end{itemize}

\subsection{Difficulty Annotation}
Difficulty is defined with respect to a \emph{competent respondent with Islamic knowledge}. Annotators should not base their judgment on personal familiarity, but instead on how difficult the question would be for this reference respondent to answer correctly. Each question is assigned a single integer score from 1 to 5 according to the definitions below.

\subsubsection{Difficulty Definitions}
\begin{table}[h]
\centering
\small
\resizebox{\columnwidth}{!}{
\begin{tabular}{c l p{0.50\columnwidth}}
\toprule
\textbf{Score} & \textbf{Label} & \textbf{Definition} \\
\midrule
1 & Very Easy & Basic factual recall, simple definitions, or straightforward yes/no questions. \\
2 & Easy & Requires basic understanding of concepts with minimal reasoning. \\
3 & Moderate & Requires understanding multiple concepts and some analytical reasoning. \\
4 & Hard & Requires deep understanding, synthesis of multiple sources/concepts, and nuanced reasoning. \\
5 & Very Hard & Requires expert-level analysis, balancing competing interests, and consideration of complex ethical frameworks. \\
\bottomrule
\end{tabular}}
\caption{Difficulty rating scale used for question annotation.}
\label{tab:difficulty_scale}
\end{table}

\subsubsection{Guidelines for Difficulty Assessment}
Annotators consider:
\begin{itemize}[noitemsep,topsep=0pt,leftmargin=*]
    \item depth of knowledge required (basic vs.\ specialized),
    \item complexity of reasoning needed (recall vs.\ application vs.\ synthesis vs.\ balancing tradeoffs),
    \item number of concepts or sources involved,
    \item level of nuance (exceptions, conditions, context sensitivity, \emph{khil\=af}),
    \item whether simple recall suffices or reasoning is necessary.
\end{itemize}
\noindent\textbf{Note:} a question can be difficult due to obscure knowledge even if it is not multi-step.

\subsubsection{Tie-break Rules}
\begin{itemize}[noitemsep,topsep=0pt,leftmargin=*]
    \item Choose the higher score if mistakes are likely due to nuance, exceptions, or competing principles.
    \item Choose the lower score if the answer is direct and reliably determined from a single well-known rule or fact.
    \item If the question is underspecified, keep the score honest and flag the issue in the interface notes (if available).
\end{itemize}

\subsection{Reasoning Assessment}
\subsubsection{\texttt{reasoning} (\texttt{true}/\texttt{false})}
\begin{itemize}[noitemsep,topsep=0pt,leftmargin=*]
    \item \texttt{reasoning = false} if the answer is simple recall/definition (no inference).
    \item \texttt{reasoning = true} if answering requires applying, interpreting, comparing, reconciling, justifying, or inferring.
\end{itemize}

\subsubsection{\texttt{multi\_step} (\texttt{true}/\texttt{false})}
Set \texttt{multi\_step = true} only if multiple logical steps/considerations are required, such as:
\begin{itemize}[noitemsep,topsep=0pt,leftmargin=*]
    \item comparing multiple sources or viewpoints,
    \item weighing competing principles (harms vs.\ benefits, conflicting obligations),
    \item applying a rule, then an exception/condition, then concluding,
    \item building a chain with intermediate conclusions.
\end{itemize}
Set \texttt{multi\_step = false} if reasoning is present but essentially one step (a single application or inference).

\subsubsection{Consistency Rules}
\begin{itemize}[noitemsep,topsep=0pt,leftmargin=*]
    \item If \texttt{reasoning = false}, then \texttt{multi\_step} must be \texttt{false}.
    \item If \texttt{multi\_step = true}, then \texttt{reasoning} must be \texttt{true}.
\end{itemize}

\subsection{Category Classification}
Annotators assign \texttt{category\_type} to exactly one of the following category names (exact strings):
\begin{itemize}[noitemsep,topsep=0pt,leftmargin=*]
    \item Islamic Creed
    \item Jurisprudence
    \item Inheritance Law
    \item Hadith Studies
    \item Qur'anic Studies
    \item Prophetic Biography
    \item Islamic History
    \item Islamic Ethics and Morality
    \item Islamic Finance and Economics
    \item Islamic Family Law
    \item Comparative Religion
    \item Contemporary Issues
\end{itemize}

\subsubsection{Boundary Rules}
\begin{itemize}[noitemsep,topsep=0pt,leftmargin=*]
    \item Modern banking/finance products $\rightarrow$ \texttt{Islamic Finance and Economics}.
    \item Marriage/divorce/custody $\rightarrow$ \texttt{Islamic Family Law}; inheritance shares/heirs $\rightarrow$ \texttt{Inheritance Law}.
    \item Hadith authentication/narrators/classification $\rightarrow$ \texttt{Hadith Studies}; hadith used mainly to derive a ruling $\rightarrow$ usually \texttt{Jurisprudence}.
    \item \emph{S\=\i rah} $\rightarrow$ \texttt{Prophetic Biography}; later eras/dynasties $\rightarrow$ \texttt{Islamic History}.
    \item Novel modern scenario $\rightarrow$ \texttt{Contemporary Issues}; timeless moral teaching $\rightarrow$ \texttt{Islamic Ethics and Morality}.
\end{itemize}

\subsection{Ambiguity and Missing Context}
Some questions may be underspecified or admit multiple valid scholarly answers. In such cases, annotators:
\begin{itemize}[noitemsep,topsep=0pt,leftmargin=*]
    \item still assign the best category based on the main domain being tested,
    \item rate difficulty based on what is required to answer responsibly (often higher if many qualifications are needed),
\end{itemize}

\begin{table}[t]
\centering
\small
\resizebox{\columnwidth}{!}{
\begin{tabular}{p{0.52\columnwidth} p{0.40\columnwidth}}
\toprule
\textbf{Question} & \textbf{Label} \\
\midrule
What is \emph{tawh\={\i}d}? &
Difficulty: 1 \newline
Reasoning: false \newline
Multi\_step: false \newline
Category\_type: Islamic Creed \\
\midrule
Explain the difference between \emph{w\=ajib} and \emph{sunnah} acts. &
Difficulty: 2 \newline
Reasoning: true \newline
Multi\_step: false \newline
Category\_type: Jurisprudence \\
\midrule
A person touched their spouse and then prayed. Does this invalidate wud\=u'? Explain. &
Difficulty: 3 \newline
Reasoning: true \newline
Multi\_step: true \newline
Category\_type: Jurisprudence \\
\midrule
Compute inheritance shares when the deceased leaves a wife, two daughters, and parents. &
Difficulty: 4 \newline
Reasoning: true \newline
Multi\_step: true \newline
Category\_type: Inheritance Law \\
\midrule
Classify a hadith given narrator reliability and continuity of isn\=ad. &
Difficulty: 4 \newline
Reasoning: true \newline
Multi\_step: true \newline
Category\_type: Hadith Studies \\
\midrule
Evaluate a modern bioethical dilemma by balancing harms/benefits and competing obligations. &
Difficulty: 5 \newline
Reasoning: true \newline
Multi\_step: true \newline
Category\_type: Contemporary Issues \\
\bottomrule
\end{tabular}}
\vspace{-0.2cm}
\caption{Examples illustrating the annotation guideline.}
\label{tab:annotation_examples}
\vspace{-0.3cm}
\end{table}
\subsection{Examples}
The examples in Table \ref{tab:annotation_examples} illustrate how to apply the labels (they are not taken from the dataset).

\subsection{Annotation Results and Agreement}
\label{app:human_valid_results}
We report summary statistics for the human annotation process and evaluate annotator consistency on a manually validated subset.

\noindent\textbf{Difficulty labels.} The difficulty annotations are reasonably distributed across the 1--5 scale, with the highest proportion at level 3 (26.90\%), followed by level 2 (23.03\%), level 4 (20.03\%), level 1 (17.67\%), and level 5 (12.37\%).

\noindent\textbf{Category assignment.} 
We observe an overall agreement rate of 82.96\% and a Cohen's $\kappa$ of 0.62, indicating substantial agreement and supporting the reliability of the annotation guidelines.

\section{Retrieval Comparison}
\label{app:retrieval-comparison}

Our main experiments focus on end-to-end answer faithfulness under grounded generation, rather than isolating the retrieval engine in a separate benchmark. To better contextualize the retrieval component used in our RAG and Agentic RAG settings, we conducted a preliminary retrieval-side comparison between our dense multilingual retriever and a sparse lexical baseline.

Specifically, we compare \texttt{mE5-base}, the dense encoder used in our retrieval-augmented experiments, against BM25 on the IslamicEval 2025 Shared Task 2 dataset \citep{mubarak-etal-2025-islamiceval}. We report Mean Average Precision at rank 5 (MAP\_Q@5). The results show that \texttt{mE5-base} outperforms BM25 by a clear margin, achieving 0.138 compared with 0.094.

This result supports our choice of dense retrieval for two reasons. \textit{First}, the task is bilingual (Arabic/English), and dense multilingual embeddings better align semantically equivalent queries across languages. \textit{Second}, Quranic question answering often depends on theological and paraphrastic phrasing that is not always well matched by pure lexical overlap. At the same time, the absolute retrieval scores remain modest, suggesting that standard RAG based retrieval is still insufficient for many questions. This observation is consistent with our main findings. Standard RAG improves correctness, but Agentic RAG yields further gains by turning retrieval into an iterative evidence-seeking process rather than a one-pass preprocessing step.

We note that this comparison serves only as a preliminary sanity check on retrieval quality, not a full benchmark. A broader study of hybrid retrieval remains for future work.

\section{Label-wise Results}
\label{app:NC_res}
Table~\ref{tab:res_NC} reports the proportions of \texttt{correct}, \texttt{incorrect}, and \texttt{not attempted} predictions in Arabic and English. Fanar-2-27B achieves the best overall performance, with an average correct rate of 48.05\%, followed by ALLaM-7B at 37.75\% and Fanar-1-9B at 35.40\%. This establishes a clear gap between the strongest models and the remaining systems.

The table also reveals different response behaviours. Models such as ALLaM-7B and Fanar-1-9B attempt most questions, but they also produce relatively high incorrect rates. By contrast, several Qwen3 and DeepSeek variants leave a large proportion of questions unanswered, often exceeding 70\% \texttt{not attempted}, which leads to low overall correctness despite fewer incorrect responses. In general, many models perform better in English than in Arabic, whereas Fanar-2-27B remains comparatively balanced across both languages. These results show that separating \texttt{correct}, \texttt{incorrect}, and \texttt{not attempted} predictions offers a more informative view of model behaviour than reporting accuracy alone.

\begin{table*}[t]
\centering
\setlength{\tabcolsep}{4pt}
\scalebox{0.75}{%
\begin{tabular}{lrrr|rrr|r}\toprule
\multirow{2}{*}{\textbf{Model}} &\multicolumn{3}{c}{\textbf{Arabic}} &\multicolumn{3}{c}{\textbf{English}} &\multirow{2}{*}{\textbf{Avg.}} \\\cmidrule{2-7}
&Correct &Incorrect &Not Attempted &Correct &Incorrect &Not Attempted & \\\midrule
\textbf{Fanar-2-27B} &48.20 &21.50 &30.30 &47.90 &30.20 &21.90 &\textbf{48.05} \\
ALLaM-7B &42.70 &52.90 &4.40 &32.80 &63.70 &3.50 &\textbf{37.75} \\
Fanar-1-9B &34.50 &54.10 &11.40 &36.30 &55.10 &8.60 &\underline{35.40} \\
AceGPT-v2-8B &23.10 &64.30 &12.60 &28.80 &57.20 &14.00 &25.95 \\
EuroLLM-9B &22.30 &67.20 &10.50 &29.10 &64.50 &6.40 &25.70 \\
SILMA-9B-v1.0 &20.40 &70.90 &8.70 &28.50 &66.10 &5.40 &24.45 \\
Qwen3-4B-2507 &15.80 &45.30 &38.90 &27.90 &45.20 &26.90 &21.85 \\
gpt-oss-20b &15.90 &22.60 &61.50 &27.20 &27.20 &45.60 &21.55 \\
Llama-3.1-8B &13.00 &74.00 &13.00 &25.80 &47.40 &26.80 &19.40 \\
Mistral-7B-v0.2 &13.50 &53.50 &33.00 &24.40 &59.10 &16.50 &18.95 \\
SeaLLM-7B-v2.5 &11.60 &76.30 &12.10 &23.80 &64.80 &11.40 &17.70 \\
Qwen2.5-3B &11.00 &61.20 &27.80 &20.00 &63.20 &16.80 &15.50 \\
Qwen3-14B &16.00 &12.50 &71.50 &14.00 &4.20 &81.80 &15.00 \\
Llama-2-7b &4.40 &47.20 &48.40 &18.80 &72.00 &9.20 &11.60 \\
DeepSeek-R1-0528-Qwen3-8B &6.30 &17.70 &76.00 &11.90 &13.30 &74.80 &9.10 \\
Qwen3-8B &8.80 &10.00 &81.20 &8.50 &5.60 &85.90 &8.65 \\
Qwen3-4B-Thinking-2507 &6.50 &3.10 &90.40 &9.40 &14.20 &76.40 &7.95 \\
Qwen3-4B &6.50 &17.30 &76.20 &9.00 &9.60 &81.40 &7.75 \\
Qwen3-1.7B &3.20 &18.10 &78.70 &5.10 &12.60 &82.30 &4.15 \\
Qwen3-0.6B &1.30 &47.00 &51.70 &5.40 &39.70 &54.90 &3.35 \\
DeepSeek-R1-Distill-Qwen-7B &1.40 &37.40 &61.20 &4.30 &37.60 &58.10 &2.85 \\
DeepSeek-R1-Distill-Qwen-1.5B &0.10 &21.60 &78.30 &1.00 &41.10 &57.90 &0.55 \\
\bottomrule
\end{tabular}
}
\caption{Results including correct, incorrect and not attempted.}
\label{tab:res_NC}
\end{table*}

\end{document}